\documentclass{article}

\usepackage{microtype}
\usepackage{amsmath}
\usepackage{graphicx}
\usepackage{subcaption}
\usepackage{booktabs} 
\usepackage{amssymb}
\usepackage{cancel}
\usepackage{fancyhdr}
\usepackage[accepted]{icml2018}

\usepackage{hyperref}

\begin{document}
\twocolumn[
\icmltitle{Predicting ice flow using machine learning}
\renewcommand{\headrulewidth}{1pt}
\pagestyle{fancy}
\fancyhead{}


\begin{icmlauthorlist}
\icmlauthor{Yimeng Min}{MILA}
\icmlauthor{S. Karthik Mukkavilli}{MILA}
\icmlauthor{Yoshua Bengio}{MILA}
\end{icmlauthorlist}

\icmlaffiliation{MILA}{Mila - Quebec AI Institute, Universit\'e de Montr\'eal, Montreal, Canada. Submitted to NeurIPS 2019 Workshop on Tackling Climate Change with Machine Learning}
\icmlcorrespondingauthor{S. Karthik Mukkavilli}{mukkavis@mila.quebec} 
\chead{\small{Neural Information Processing Systems (NeurIPS) 2019, Workshop on Tackling Climate Change with Machine Learning}}



\vskip 0.3in
]



\printAffiliationsAndNotice{}  

\begin{abstract}
  Though machine learning has achieved notable success in modeling sequential and spatial data for speech recognition and in computer vision, applications to remote sensing and climate science problems are seldom considered. In this paper, we demonstrate techniques from unsupervised learning of future video frame prediction, to increase the accuracy of ice flow tracking in multi-spectral satellite images. As the volume of cryosphere data increases in coming years, this is an interesting and important opportunity for machine learning to address a global challenge for climate change, risk management from floods, and conserving freshwater resources. Future frame prediction of ice melt and tracking the optical flow of ice dynamics presents modeling difficulties, due to uncertainties in global temperature increase, changing precipitation patterns, occlusion from cloud cover, rapid melting and glacier retreat due to black carbon aerosol deposition, from wildfires or human fossil emissions. We show the adversarial learning method helps improve the accuracy of tracking the optical flow of ice dynamics compared to existing methods in climate science. We present a dataset, IceNet, to encourage machine learning research and to help facilitate further applications in the areas of cryospheric science and climate change.
\end{abstract}

\section{Introduction}

Recent developments in the climate sciences, satellite remote sensing and high performance computing are enabling new advancements that can leverage the latest  machine learning techniques. Petabytes of data is being produced from a new-generation of Earth observation satellites, and the commercialization of the space industry, is driving costs of acquiring data down further. Such large geoscience datasets, coupled with latest supercomputer simulation outputs from global climate model intercomparison projects is now available for machine learning applications \citep{Kay2015,Schneider2017, Reichstein2019}. 

In another related study, \cite{ogorman2018} used random forests to parameterize moist convection processes to successfully emulate physical processes from expensive climate model outputs. \cite{scher2018} was also able to approximate the dynamics of a simple climate model faithfully after being presented enough data with deep learning. 
    
Recently machine learning has demonstrated promise in resolving the largest source of uncertainty \citep{Sherwood2014,ipcc_global_2018} in climate projections, cloud convection \citep{Rasp2018, Gentine2018}. \cite{Rasp2018} and \cite{Gentine2018} demonstrated the use of deep learning in emulating sub-grid processes to resolve model clouds within simplified climate models at a fraction of the computational cost of high resolution physics models. These developments provide machine learning researchers opportunities to build models about ice flow and dynamics from satellite data, or using new video prediction techniques \citep{Mathieu2016,denton2018stochastic} to predict changes in glaciers and ice dynamics.

In this work, our main contributions are (1) development of an unsupervised learning model to track ice sheet and glacier dynamics; and (2) introducing IceNet, a dataset that we make available for the community to serve as a first step in bridging gaps between machine learning and cryosphere climate research.

\section{Dataset}

\label{headings}
\begin{figure}[htbp]
  \centering
    \includegraphics[width=6cm]{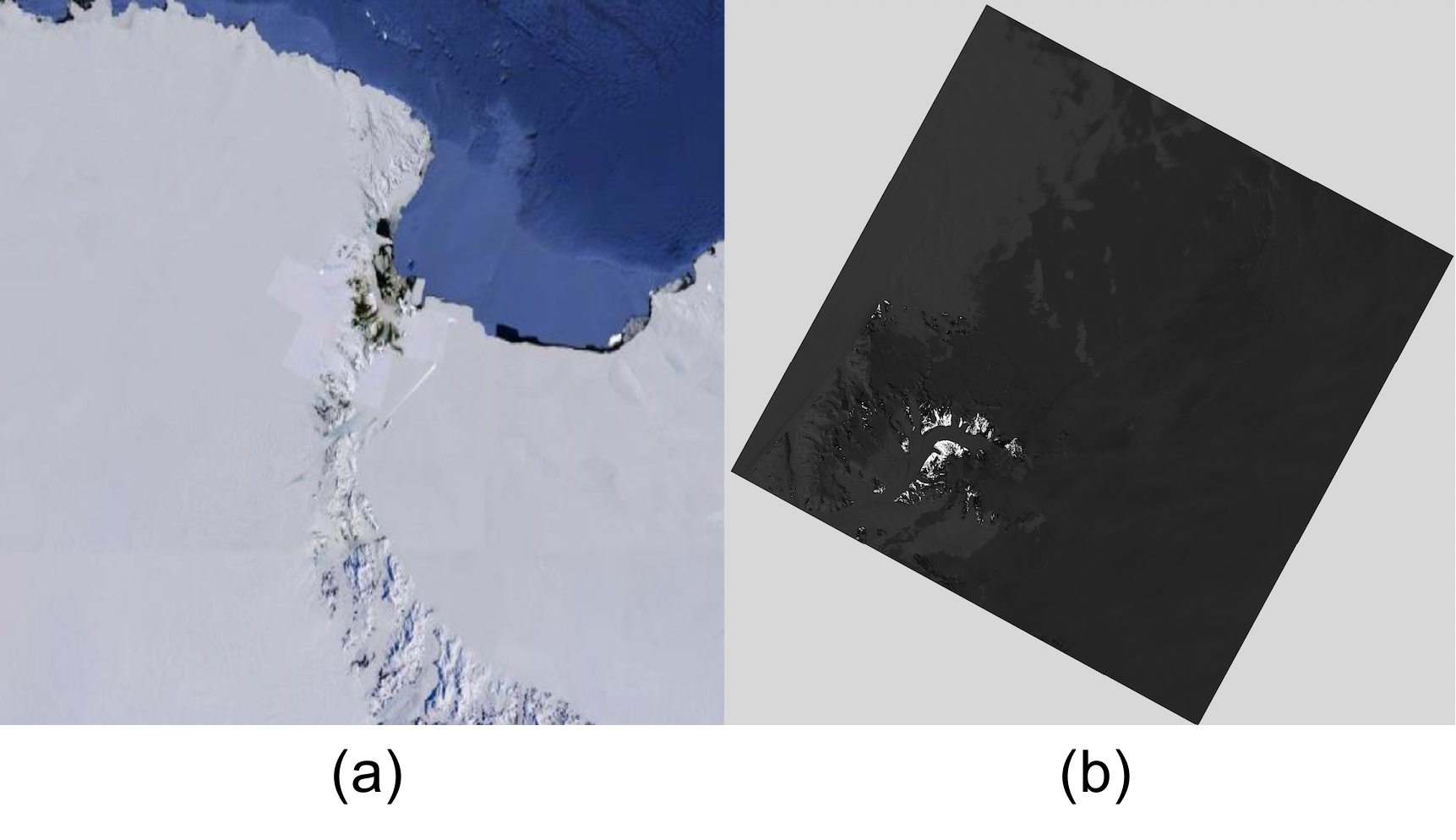}

  \caption{(a) The region our dataset investigates. (b) Coastal signal(Band 1, 0.43$\mu$m to 0.45 $\mu$m) collected by the LANDSAT 8 at 2015 November 22. The four corners contain no information.}
  \label{fig:antar}
\end{figure}

In this paper, we investigate on seven bands ranging from 0.43 $\mu$m to 2.29$\mu$m(visible, near-infrared and shortwave light) with a resolution of 30 meters.  The details of LANDSAT 8 can be found at https://www.usgs.gov/land-resources/nli/landsat.  In our dataset, we focus on a particular area at Antarctica with a latitude of 80$^{\circ}$01'25''  South and longitude 153$^{\circ}$11'10'' East, where the ice flow's moving pattern is dominated by the Byrd Glacier (path 54, row 118 using worldwide reference system-2).  
 The picture is shown in Figure \ref{fig:antar}.

    

Our dataset contains the satellite images ranging from November 2015 to February 2017, with total 10675 images and every image has 12 frames with the shape of 128 by 128; the interval between each frame ranges from two weeks to 9 month gaps, each pixel stands for a 30 meters by 30 meters region.
\subsection{Labels}
The images are denoted as ${F_i}$ where $i$ is from 1 to 12 and the frames(subscenes) in each image are $x^j_i \in R^{128 \times 128}$, where $i\in \{1...12\}$ and $j \in \{1...1525\}$. For finding the next subscene, or chip, that matches the $x^j_{i-1}$  best, we compare the $x^j_{i-1}$ to a range of possible regions by calculating the correlation between two chips, the equation writes as:
\begin{equation}
CI(r,s) =\frac{\sum_{mn}(r_{mn}-\mu_r)(s_{mn} - \mu_s)}{[\sum_{mn}(r_{mn}-\mu_r)^2]^{1/2}[\sum_{mn}(s_{mn} - \mu_s)^2]^{1/2}}
\end{equation}
where r and s are the two images and $\mu$ is the mean value.
\begin{figure}[h]
  \centering
    \includegraphics[width=0.5\textwidth]{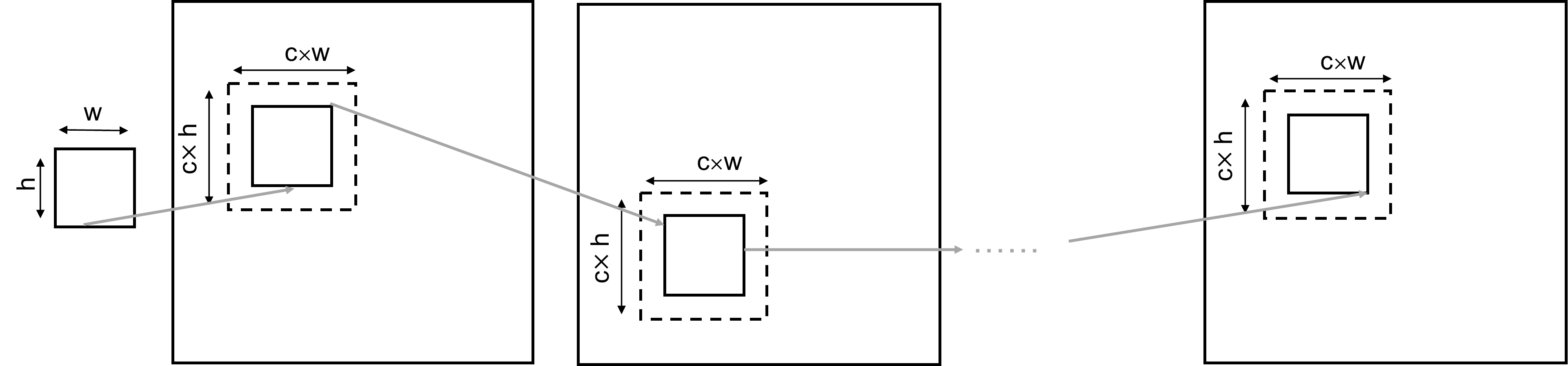}

  \caption{A larger subscene is selected in case of the previous subscene moving outside the original grid.}
  \label{fig:mov}
\end{figure}
The ice flow is not static, moving areas of the large ice sheets remain a challenge for tracing the ice flow. For tackling surface feature movement, we select a larger area by a scale factor $c(c>1)$ that centres around the previous subscene in case the pattern moving outside the previous grid,  the most correlated one is chosen as the next subscene(the ground truth). The pipeline is shown in Figure~\ref{fig:mov}. 









\section{Model}
We use a stochastic video generation with prior for prediction. 
The prior network observes frames $\textbf{x}_{1:t-1}$ and output $\mu_{\psi}(\textbf{x}_{1:t-1})$ and $\sigma_{\psi}(\textbf{x}_{1:t-1})$ of a normal distribution and is trained with by maxing:
\begin{equation}
\begin{aligned}
\mathcal{L}_{\theta,\phi,\psi}(x_{1:T}) = \sum_{t=1}^T \big{[} \mathbb{E}_{q_{\phi}(z_{1:t} | x_{1:t})} log p_{\theta}(x_t | x_{1:t-1}, z_{1:t}) \\
 - \beta D_{KL}(q_{\phi}(z_t | x_{1:t}) || p_{\psi}(z_t | x_{1:t-1}))\big{]}
\end{aligned}
\end{equation}
Where $p_{\theta}$, $q_{\phi}$ and $p_{\psi}$ are generated from convolutional LSTM. $q_{\phi}$ and $p_{\psi}$ denote the normal distribution draw from $\textbf{x}_t$ and $\textbf{x}_{t-1}$ and  $p_{\theta}$ is generated from encoding the $\textbf{x}_{t-1}$ together with the $\textbf{z}_t$.

Subscene $\hat{\textbf{x}}_t$ is generated from a decoder with a deep convolutional GAN architecture a by sampling on a prior $\textbf{z}_t$ from the latent space drawing from the previous subscenes combined with the last subscene $\textbf{x}_{t-1}$. After decoding, the predict subscene is passed back to the input of the prediction model and the prior.
The latent space $\textbf{z}_t$ is draw from $p_\psi(\textbf{z}_t|\textbf{x}_{1:t-1})$. 

The loss of our model contains three parts, KL divergence of the prior loss $D_{KL}$, a $\ell_2$ penalty between $\hat{\textbf{x}}_t$ and $\textbf{x}_t$ and an additional $\ell_2$ penalty of the area centred around the peak of every subscene. The prediction results vary with different weight of $\ell_2$ penalty on the peak, when the weight is too small, the model may ignore the low frequency of the subscene and $\hat{\textbf{x}}_t$ will predict the noisy small textures(crevasses) of the ice flow corresponding to $\hat{\textbf{x}}_t$. When increasing the weight, the model predicts the peak regions but fails to generate the small textures of the ice flow.
\section{Experiment Results and Discussion}
\begin{figure}[t]
  \centering
    \includegraphics[width=0.5\textwidth]{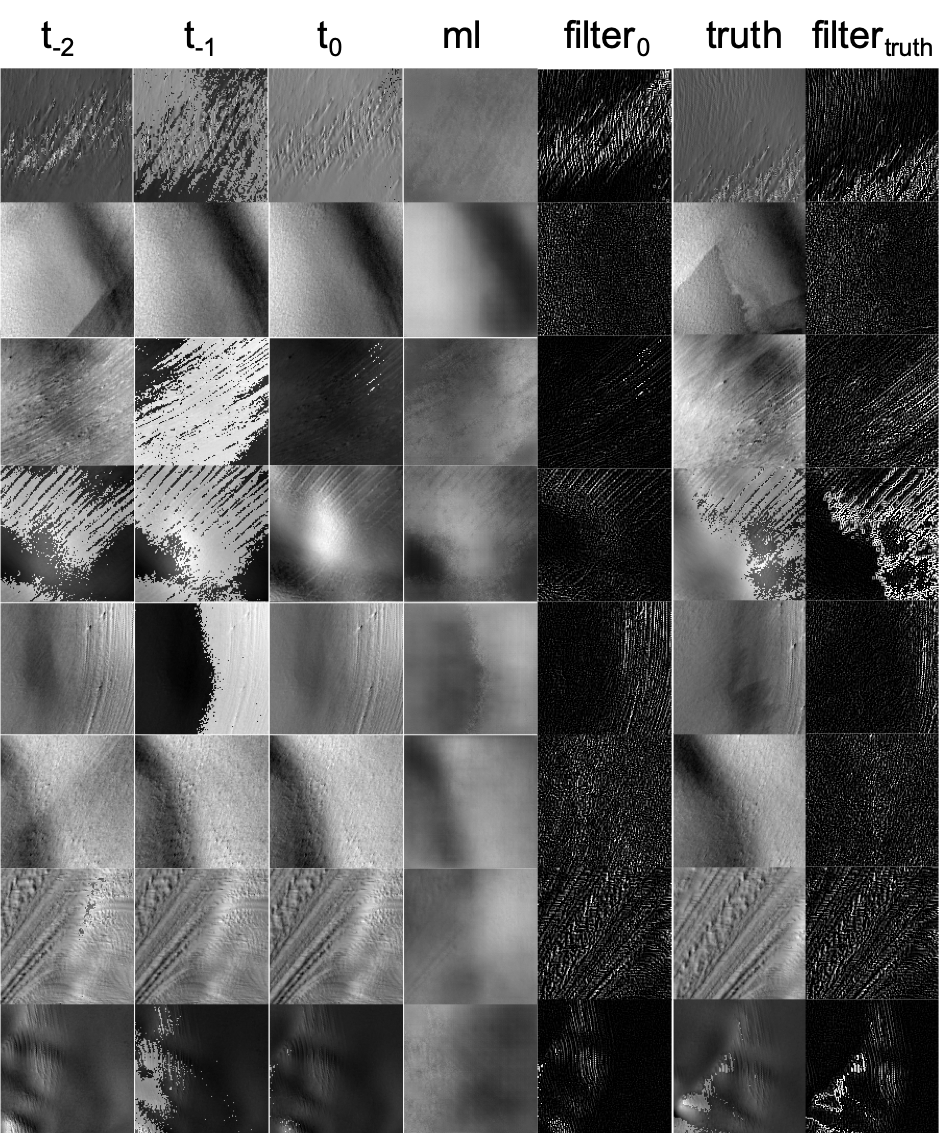}
  \caption{Subscenes generated with different models, the first three columns: the past three subscenes; the fourth column: machine learning predicted next subscene; fifth column: high pass of $t_0$; sixth and seventh column: the subscene on $t_1$ and $t_2$; eighth column: high pass of $t_2$.}
  \label{fig:model}
\end{figure}

\begin{figure}[h]

  \centering
    \includegraphics[width=0.5\textwidth]{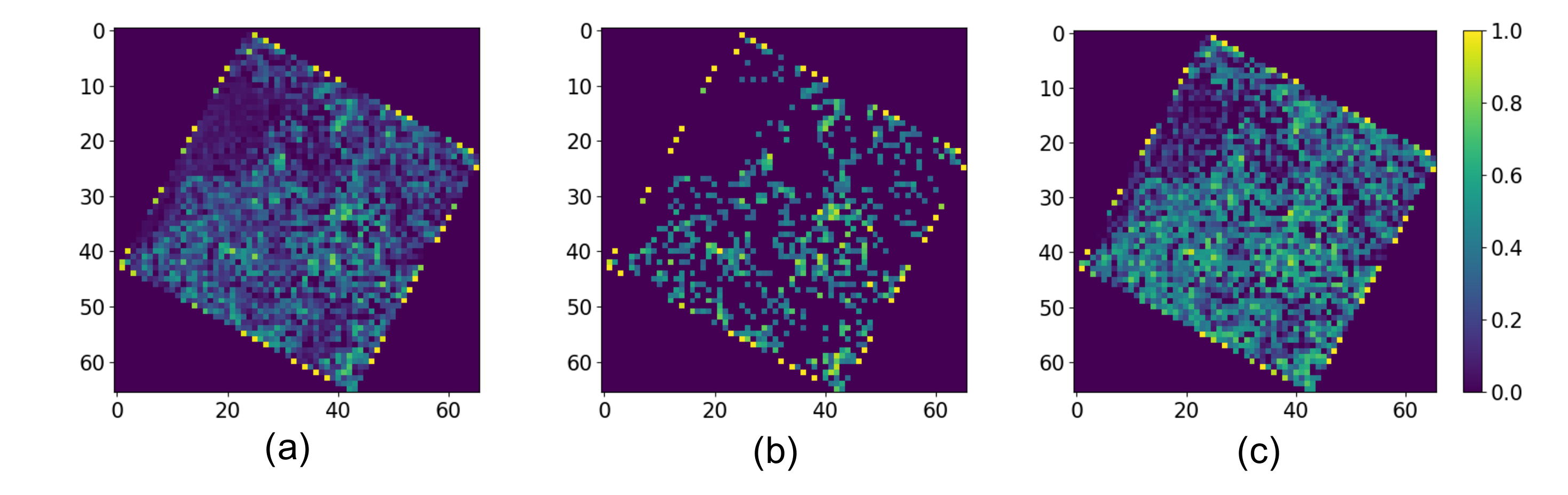}
  \caption{The correlation map. a) persistence model(correlation between $t_0$ and $t_2$); b) high frequency model (correlation between filter$_0$ and filter$_2$); c) machine learning model(correlation between ml and $t_2$).}
  \label{fig:corr}
\end{figure}

We train our model with $\textbf{z} \in R^{128}$ and 2 LSTM layers, each layer has 128 units. By conditioning on the past eight subscenes, the results of our model on different types of subscenes are shown in Figure ~\ref{fig:model} and \ref{fig:corr}. 
For ice flow pattern with proper slopes(not too steep), e.g. line 2 and 6 in Figure 4, the machine learning can reproduce the slopes’ shapes and positions, resulting in successfully correlating two subscenes. In the experiment, the capability of reproducing small textures grows as enlarging the hidden space and batch size. However, the high pass filter's performance differs in this two examples: in line 2, the high pass model draws the textures from $t_0$ and $t_2$, since the high pass filter's results are close to binary, as long as the textures are extracted, two subscenes correlate. However, for line 6, the filter on $t_0$ generates noisy signals, resulting in the failure of correlating. Another example the high pass filter fails is line 3, when the previous $t_{0}$ does not collect the texture information(the satellite signal is affected by the cloud), in this case, the filtered subscene lacks the key information to be correlated with the filter$_2$.

The machine learning model avoids the poisonous $t_{0}$ by generating parameters learned from a range of past subscenes. In this case, though some of the past subscenes' signals are contaminated, the model can successfully reproduce the slope and small patterns, as shown in line 3 in Figure ~\ref{fig:model}. By adding proper weight around the peak area, the model successful reproduces the peak and learns the small textures from previous subscenes, as shown in line 3 and 4. The model also generates a continuous range of pixels that help reduce the correlation error, which is different from the binary result generating from the high pass filter. For flat regions with complex textures, e.g. line 1 and line 7, the persistence model correlates when both $t_0$ and $t_2$ parse the patterns(not affected by the cloud). The overall correlation map is shown in Figure~\ref{fig:corr} and the statistical results are shown in Table~\ref{table2}. The machine learning model helps improve the overall mean correlation\footnote{Mean correlation is generated from the non-zeros} comparing with persistence model and high pass filter model. For some flat regions with clear pattern like crevasses, an example can be found at line 7 in Figure ~\ref{fig:model}, the high pass filter correlates better. However, these kinds of regions count for a small percentage in the area we investigate, resulting in the improvement in high correlation subscenes using the high pass filter while overall worse performance due to the noisy binary pixels. The machine learning model enlarges the medium correlation regions by generating continuous pixels, the peak area and learning from a range of past frames instead of just $t_0$. 







\begin{table}[h]
\tiny
  \centering
  \begin{tabular}{p{1cm}p{0.6cm}p{0.3cm}p{0.3cm}p{0.3cm}}
    \toprule
    \multicolumn{1}{c}{} & \multicolumn{1}{c}{ } & \multicolumn{1}{c}{Persistence(Last frame)} &\multicolumn{1}{c}{Hi-pass Filter}&\multicolumn{1}{c}{Machine learning}                  \\ 
    \cmidrule(r){3-5}
    Correlation&  Mean    & 0.237     & 0.201 &0.362 \\
    \midrule
    Low & $<$ 0.3&  0.699 & 0.598  & 0.393     \\
     Medium &0.3$\sim$0.7&0.271     & 0.337 & 0.557     \\
    High &$>$ 0.7& 0.0300     & 0.0651       & 0.0504  \\
    \bottomrule
  \end{tabular}
    \caption{Results of three models}
  \label{table2}
\end{table}


\section{Conclusions and Future Work}
We present IceNet dataset and encourage machine learning community to pay more attention to socially and scientifically relevant datasets in the cryosphere and develop new models to help combat climate change. We also use an unsupervised learning model to predict future ice flow. Comparing to the high pass filter or persistence model, our model correlates the past and present ice flow better. Our model can also be improved if more physical and environmental parameters are introduced into the model, for example, the wind speed and the aerosol optical depth components in the atmosphere. The first parameter provides a trend for the ice flow movement and the second parameter gives us a confidence factor about the satellite images' quality, dropout to particular frames can be applied if the aerosol optical depth rises over a threshold. Furthermore, black carbon aerosols were found to accelerate ice loss and glacier retreat in the Himalayas and Arctic from both wildfire soot deposition and fossil fuel emissions. Detailed analysis of the feedback effects in 'black ice' would be a future avenue of research 

The images of IceNet dataset is very different from traditional video datasets, such as in the moving-mnist, some areas are dominated by 'small textures' while some can be smooth areas with peaks. This suggests that the transfer capability of existing models need to be investigated further or new models need to be developed for predicting the ice flow on different types of terrains around the planet.








\bibliography{ref} 
\bibliographystyle{icml2018}
\end{document}